\documentclass[sigconf]{acmart}
\usepackage{amsmath,amsfonts}
\usepackage{algorithmic}
\usepackage{algorithm}
\usepackage{graphicx}
\usepackage{textcomp}
\usepackage{xcolor}
\usepackage{microtype}
\usepackage{tabularx}
\usepackage{graphicx}
\usepackage{booktabs}
\def\BibTeX{{\rm B\kern-.05em{\sc i\kern-.025em b}\kern-.08em
    T\kern-.1667em\lower.7ex\hbox{E}\kern-.125emX}}

\begin{document}
\fancyhead{}

\title{SparseDNN: Fast Sparse Deep Learning Inference on CPUs}

\author{Ziheng Wang}
\email{zihengw@stanford.edu}
\affiliation{%
  \institution{Stanford University}
}

\begin{abstract}
The last few years have seen gigantic leaps in algorithms and systems to support efficient deep learning inference. Pruning and quantization algorithms can now consistently compress neural networks by an order of magnitude. For a compressed neural network, a multitude of inference frameworks have been designed to maximize the performance of the target hardware. While we find mature support for quantized neural networks in production frameworks such as OpenVINO and MNN, support for pruned sparse neural networks is still lacking. To tackle this challenge, we present SparseDNN, a sparse deep learning inference engine targeting CPUs. We present both kernel-level optimizations with a sparse code generator to accelerate sparse operators and novel network-level optimizations catering to sparse networks. We show that our sparse code generator can achieve significant speedups over state-of-the-art sparse and dense libraries. On end-to-end benchmarks such as Huggingface pruneBERT, SparseDNN achieves up to 5x throughput improvement over dense inference with state-of-the-art OpenVINO. Open source library at https://github.com/marsupialtail/sparsednn
\end{abstract}
\maketitle




\vskip 0.3in




\section{Introduction}

Neural networks have started to top benchmarks in computer vision (CV) and natural language processing (NLP). However, recent neural network architectures, such as deep convolutional networks and transformer networks like BERT, suffer from high memory footprint and FLOP count due to the massive number of weights \cite{devlin2018bert,howard2017mobilenets}. For example, while GPT-3 can accomplish an array of amazing feats, it has a whopping 175 billion parameters \cite{brown2020language}. Assuming the model is in 32-bit floating point, it would require around 20 V100 GPUs just to store the model. 

To facilitate model deployment in production systems with strict memory and latency constraints, there has been significant research interest in how to compress deep neural networks (DNN) \cite{gale2019state,han2015deep}. The objective is to significantly reduce the number of parameters in a DNN to reduce the inference latency and memory footprint, while not losing too much accuracy. Recent neural network pruning research has established that different criteria, such as L0 norm or L1 norm (magnitude), can be used to remove as much as 90\% of the weights in modern neural networks with little loss in accuracy \cite{molchanov2017variational,louizos2017learning,gale2019state}, if there are no constraints on which weights can be removed. These unconstrained pruning techniques typically result in unstructured sparse weight matrices.

Unfortunately, it was soon recognized that unstructured sparsity patterns are hard to support efficiently on modern CPUs and GPUs typically used for deep learning inference. This spurred the developments of custom hardware \cite{han2016eie} as well as ``structured'' pruning methods that prune blocks of weights at once. The resulting weights from these structured pruning methods often can be used directly in dense BLAS routines or have enough structure to support highly performant implementations on current hardware \cite{gray2017gpu,yao2019balanced}. However, these structured pruning methods often lead to larger accuracy losses than unstructured pruning and often do no better than a smaller dense network \cite{closerlook2019}.

In this work, we present an inference engine, SparseDNN, to efficiently support unstructured sparse deep learning inference. We show that through both network-level and kernel-level optimizations, we can significantly speed up the execution of unstructured sparse neural networks on modern datacenter CPUs where most deep learning inference takes place today \cite{mckinsey}. The design of SparseDNN is inspired by other state-of-the-art inference systems for dense neural networks such as TensorRT, OpenVINO and MNN \cite{tensorrt,jiang2020mnn,gorbachev2019openvino}, as well as performance engineering research targeting sparse matrix multiplications and convolutions in Intel Libxsmm and SkimCaffe \cite{georganas2019high,park2016faster}. SparseDNN leverages both novel network-level optimizations such as weight permutation and kernel-level optimizations through a sparse kernel generator which achieves significantly superior performance than Libxsmm and SkimCaffe \cite{georganas2019high,park2016faster}. End-to-end, we demonstrate up to 5x speedup over dense inference with OpenVINO on sparse networks like Huggingface pruneBERT \cite{sanh2020movement}.

\section{Background}
This section introduces some background concepts that motivate the development of SparseDNN.

\subsection{Deep Learning Inference Engines}
As neural network models become increasingly commonplace in production, there have been a number of different open and closed source inference frameworks from both academia and industry. Some notable examples include OpenVINO for CPUs, TensorRT for Nvidia GPUs and MNN for mobile devices \cite{gorbachev2019openvino,tensorrt,jiang2020mnn}. Almost all deep learning inference engines consist of two stages: preparation and execution. In the preparation stage, optimizations are performed on the neural network to generate a binary suitable for execution. Typically, this only has to be done once for each network and hardware backend, which allows significant computational resources to be spent as they can be amortized over all subsequent inference requests in the execution stage. The preparation stage typically determines how efficiently a neural network can be executed.

Most of the deep learning inference engines divide the preparation stage into two phases. In the first phase, network level optimizations such as layer fusion, removing redundant operators and data format selection, are performed. Latest neural network inference engines such as OpenVINO and TensorRT have become sophisticated enough to perform fusions of adjacent depthwise and groupwise convolutions \cite{gorbachev2019openvino,tensorrt}. Many of these optimizations, such as layer fusion, transfer directly to sparse neural networks. In addition, SparseDNN leverages network-level optimizations specific to sparse neural networks, such as weight permutation.

In the second phase, each operator in the optimized network is then mapped to an efficient kernel implementation depending on the target hardware. The kernels can be selected from a kernel library or generated specifically for the operator. OpenVINO and TensorRT rely on Intel oneDNN (formerly MKL-DNN) and Nvidia cuDNN respectively \cite{gorbachev2019openvino,tensorrt}. MNN relies on a semi-automated search technique to generate the kernels from a pre-defined number of optimization strategies \cite{jiang2020mnn}. TVM takes it a step further and performs compilation and autotuning for each kernel \cite{chen2018tvm}. In SparseDNN, we adopt the last approach. For each sparse operator, we use a sparse code generator to compile a kernel optimized for its sparsity pattern. 

Crucially, there is often a feedback loop between the first phase and the second phase. For example, if the kernel library's version of fused depthwise-groupwise convolution is slower than doing them separately, the fusion should be rolled back in the first phase. In the case of SparseDNN, there are sparsity patterns that our sparse code generator prefers, which guides the weight permutation optimizations in the first phase. 

Once the neural network is primed for execution, there are typically two modes of inference, synchronous and asynchronous. On CPU inference engines such as OpenVINO, synchronous inference uses all available CPU cores to process one input example at a time through the neural network to minimize latency, whereas asynchronous inference allows each CPU to process a different stream of input examples to avoid synchronization costs to maximize throughput \cite{gorbachev2019openvino}. 

\begin{figure*}[t]
\begin{center}
\includegraphics[width=0.9\linewidth]{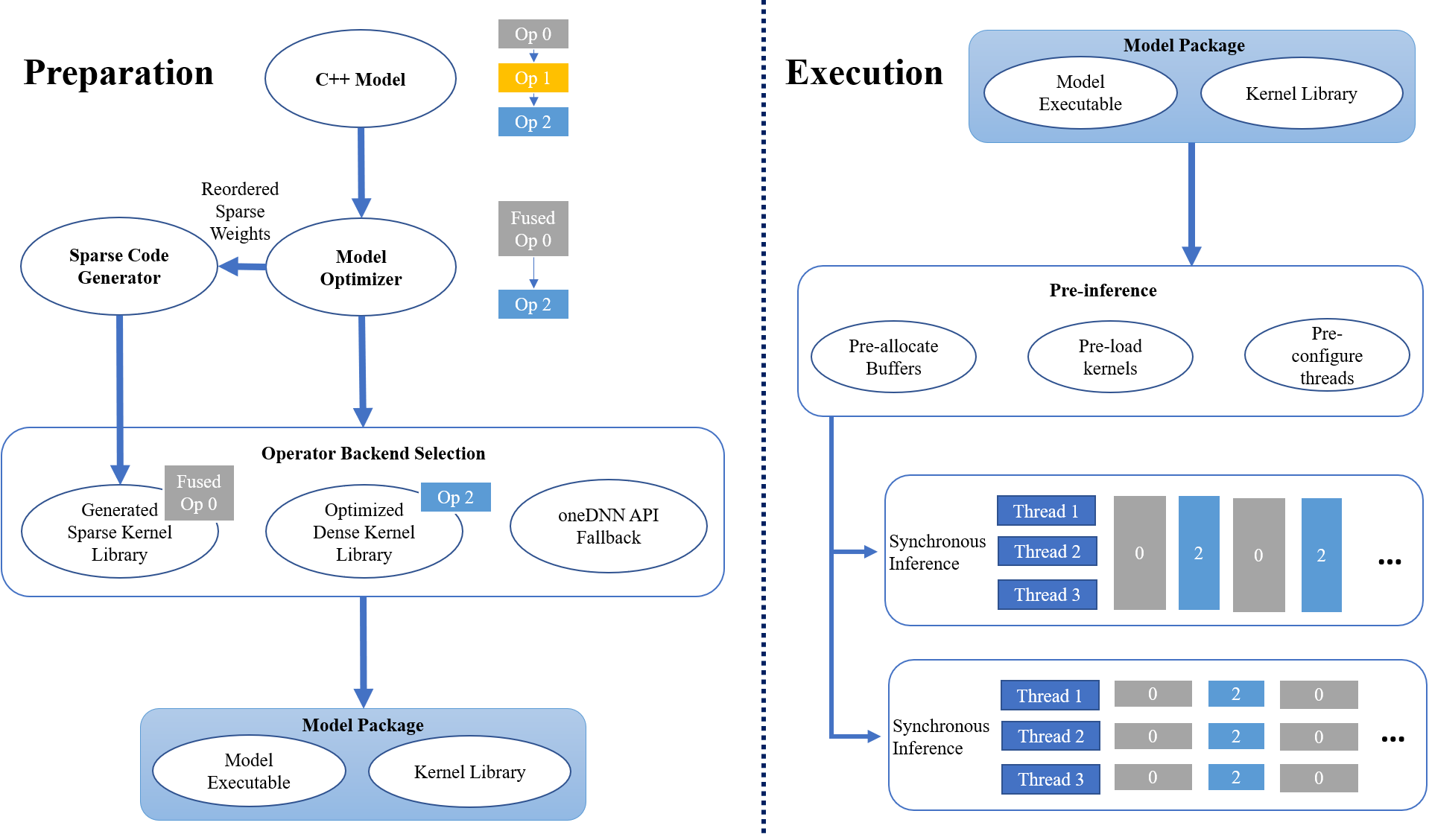}
\end{center}
   \caption{The architecture of the SparseDNN system. The preparation phase is shown on the left while the execution phase is shown on the right.}
\label{fig:arch}
\end{figure*}

\subsection{SpMM and Sparse Convolutions}
It is evident that the quality of the kernel implementations of the operators is a crucial factor in the performance of the inference engine. Dense deep learning libraries typically rely on highly optimized BLAS libraries such as oneDNN and cuDNN that builds on decades of performance engineering research \cite{goto2008high,chetlur2014cudnn}. 

In a sparse neural network in CV or NLP, the core operation at inference time is typically \textbf{sparse matrix -- dense matrix multiplication} (abbreviated as \textbf{SpMM}) and sparse convolution \cite{elsen2019fast,howard2017mobilenets,tan2019efficientnet, devlin2018bert,wang2019structured}. Sparse convolution can in turn be cast as a SpMM computation either through the im2col transformation \cite{chetlur2014cudnn} or direct convolution \cite{park2016faster}. This suggests that speeding up SpMM is critical in speeding up sparse inference. 

Although SpMM is a highly optimized kernel in scientific computing, we find that many existing SpMM implementations are poorly suited for accelerating unstructured sparse DNNs \cite{yang2018design,hong2019adaptive,hong2018efficient}. They are often catered to scientific computing applications where the sparse matrix is massive and highly sparse. A typical sparse matrix in a SpMM problem, circuit5M \cite{davis2011university}, is 5.5 million by 5.5 million with less than 0.01\% nonzeros. Unfortunately, sparse matrices encountered in deep learning are small and only moderately sparse. A typical sparse matrix in a deep learning problem, the sixth layer in a pruned MobileNet V1 network, is 512 by 512 with 10\% nonzeros \cite{elsen2019fast}. 

Many recent works have attempted to cater to SpMM problems that arise in deep learning. Intel Libxsmm and SkimCaffe provide efficient sparsity support tailored for small matrix sizes encountered in deep learning on CPUs \cite{georganas2019high, park2016faster}. We use these as the main benchmark for SparseDNN's kernels in this paper. In addition, Elsen et. al demonstrated that unstructured sparsity can be fast on Arm mobile processors with sufficient performance engineering \cite{elsen2019fast}. Sputnik and SparseRT demonstrated similar results on Nvidia GPUs \cite{gale2020sparse,wang2020sparsert}. 

The approach of SparseRT's GPU optimizations is particularly interesting. The author recognizes that the sparsity patterns of the sparse weight matrices in deep learning are statically known before inference. The sparsity pattern can thus be leveraged to optimize kernel performance through the inspector-executor approach \cite{strout2004sparse}. SparseDNN also incorporates this technique to optimize CPU SpMM kernels. In addition, we take it one step further through incorporating network-level optimizations that statically reorder the sparsity patterns in the sparse weights guided by the preferences of our code generator. 

\section{SparseDNN Architecture}
This section describes the architecture of the SparseDNN sparse deep learning inference system. The overall architecture, illustrated in Figure \ref{fig:arch}, is similar to other inference systems such as OpenVINO, TensorRT and MNN \cite{jiang2020mnn,tensorrt,gorbachev2019openvino}. 

To set up a neural network for inference in SparseRT, the neural network is first implemented using SparseDNN's C++ API. We selected this approach instead of parsing from a higher-level representation in Tensorflow or Pytorch because of its flexibility. Existing high-level frameworks also only typically have incomplete and experimental support for sparse operators. 

SparseDNN adopts the aforementioned preparation-execution paradigm. During the preparation stage, SparseDNN performs network level optimizations, and selects the best kernel implementation for each layer. SparseDNN also allows for both synchronous and asynchronous inference in the execution stage. 

\subsection{Network Optimizations}
In SparseDNN, extensive network-level optimizations are performed in the preparation phase. SparseDNN supports standard layer fusion strategies such as fusing a convolution layer and a subsequent element-wise layer. SparseDNN also finds the best kernel implementation for each operator in the neural network. Sparse kernels are generated from a sparse code generator based on their associated sparse weight matrix, which is assumed to be statically known. The sparse code generator will be discussed in detail in the next subsection. For dense kernels, SparseDNN selects between a default Intel oneDNN backend and SparseDNN's own handtuned kernel library, which can be more performant in certain cases. 

In addition to these standard optimizations performed in the prepration phase, SparseDNN also performs network-level optimizations specific to sparse neural networks. We will describe two such optimizations below. 

\subsubsection{Activation Buffer Reuse}
Memory buffers that store input activations for earlier layers of the network can be repurposed and rewritten in later layers. This is a simple optimization that is often done to minimize the memory footprint of the inference engine on edge devices \cite{jiang2020mnn}. However, on sparse networks, it also has a big impact on inference performance. Sparse operators are typically memory-bound instead of compute-bound. As a result, their performance can often be substantially improved if their inputs and outputs physically reside in cache rather than DRAM. On CPUs, the more frequently used a memory buffer is, the more likely it resides in cache. As a result, cache hit rates can be improved with activation buffer reuse.

\subsubsection{Weights Permutation}
This is an optimization that is only relevant to sparse networks. For dense linear algebra operations such as matrix multiplication, permuting the rows of the dense matrix has no effect on the operator's runtime. However, for sparse operators, permuting the rows of the sparse weight matrix can substantially change the operator's performance by changing the sparsity pattern. There have been many recent works in the scientific computing community that examine beneficial reorderings to apply to a sparse matrix \cite{jiang2020novel}.

In the context of neural network inference, permuting the rows of a sparse weight matrix by itself is unacceptable as it will lead to the wrong output. However, if a sparse matrix multiplication is followed by another sparse matrix multiplication for example, one could permute the \textit{columns} of the second sparse matrix to offset the permutation of the rows of the first sparse matrix and preserve the same output. This observation allows us to change some of the sparsity patterns of the weight matrices in the network in the preparation phase.

In the results presented here, we apply a simple greedy algorithm, shown in Algorithm \ref{alg:spmm}, to permute the rows of the weight matrices to maximize the similarity between the nonzero patterns between adjacent matrices. First, the row with the most number of nonzeros is selected to be the new first row. Then, the row with the most number of nonzero columns which overlapped with the nonzero columns in the previous row is selected to be the next row. This process is repeated until there are no rows left. We empirically found that this simple approach performed comparably to more complicated approaches that use locality sensitive hashing and clustering on the problems tested in this paper \cite{jiang2020mnn}. It is straightforward to experiment with different algorithms for different neural networks with SparseDNN.

\begin{algorithm}[tb]
   \caption{Weight Reordering Algorithm, $nnz(A[i])$ stands for the set of nonzero column positions in the $i^{th}$ row of A.}
   \label{alg:spmm}
\begin{algorithmic}[1]
   \STATE {\bfseries Input:} sparse matrix $A$ with shape $m \times n$
   \STATE {\bfseries Output:} permuted sparse matrix $B$ with shape $m \times n$
   \STATE $rem = \{0 ... m\}$
   \STATE $row = \underset{x \in rem}{\arg\max} (|nnz(A[x])|)$
   \STATE $B[0] = A[row]$ 
   \STATE $rem$.remove($row$)
   \FOR{$i$ in $1 ... m$}
    \STATE $row = \underset{x \in rem}{\arg\max}$ $(|  nnz(B[i-1]) \cap nnz(A[x])|)$
    \STATE $B[i] = A[row]$ 
   \STATE $rem$.remove($row$)
   \ENDFOR
\end{algorithmic}
\end{algorithm}

\subsection{Sparse Code Generator}
After network optimizations, SparseRT attempts to map each operator in the network to an optimized kernel implementation. Kernels for sparse operators are generated from a sparse code generator at this stage. We recognize that most operators in deep learning can be represented at their core by a matrix multiplication \cite{georganas2019high}. The sparse code generator focuses on generating high performance SpMM routines. We support sparse convolutions by converting them into a SpMM problem with the direct convolution method described in \cite{park2016faster}. 

\subsubsection{High Performance Dense GeMM}
The optimization strategies employed by our sparse code generator are inspired by optimization techniques commonly used in dense linear algebra, in particular, dense matrix multiplication (GeMM). We briefly review these techniques here for reference. 

\begin{figure}[t]
\begin{center}
\includegraphics[width=1\linewidth]{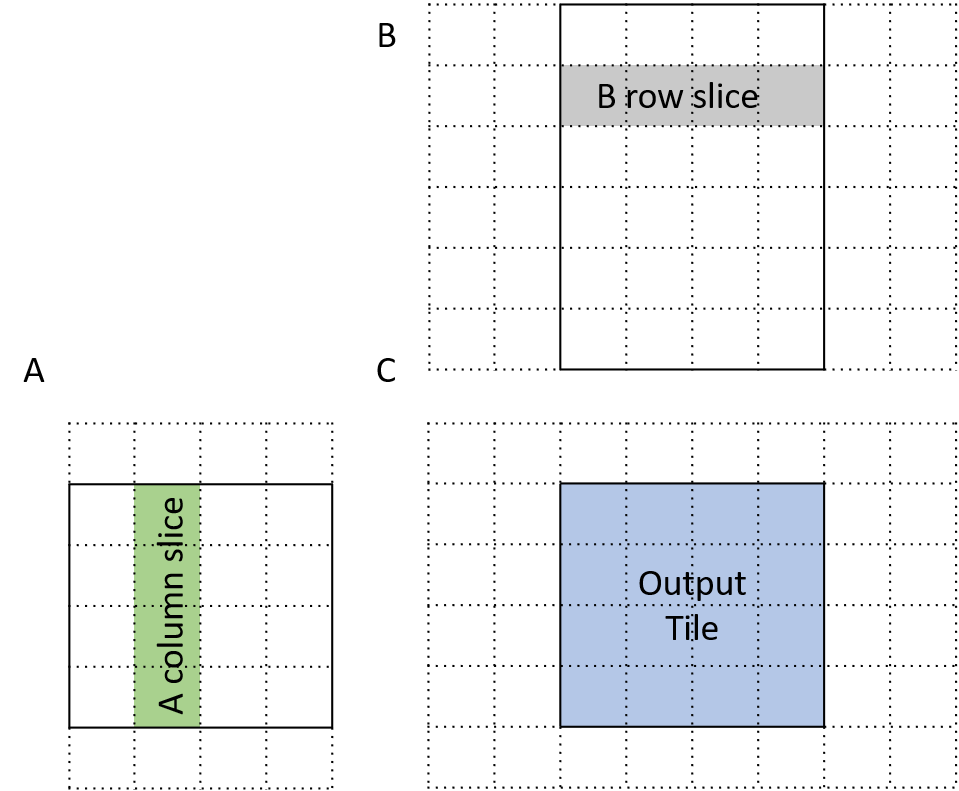}
\end{center}
   \caption{Strategy for GeMM. An output tile in C is produced by multiplying two input tiles from A and B. The microkernel computes the outer product between a column in the A tile and a row in the B tile. This outer product is accumulated to the output tile. The program then calls the microkernel again on the next column in A and the next row in B.}
\label{fig:gemm}
\end{figure}

High performance GeMM kernels typically start with a highly tuned microkernel which performs a small matrix multiplication using the SIMD instructions of the CPU. For example, a typical strategy for this microkernel is illustrated in Figure \ref{fig:gemm}. The microkernel computes the outer product between a column slice in A and a row slice in B. It typically employs register blocking to make the best use of the vector registers. By tuning the size of the A and B slices, it can trade off the number of loads vs. the number of computes to strike the best load/compute ratio for a particular hardware architecture. 

Armed with this tunable microkernel, the GeMM problem becomes how to decompose the computation into a sequence of microkernel calls. Typically, the implementation then optimizes for memory locality by using tiling. Tiling breaks down the matrix multiplication into chunks that are small enough to fit into on-chip caches on the CPU, which often offer an order of magnitude faster access than the off-chip main memory. Typically, a tile from each input matrix operand is loaded, and a tile of the output is computed using the microkernel.

We will now proceed to describe our SpMM strategy, which applies the key aforementioned optimization strategies of vectorization, register blocking and tiling. 

\subsubsection{SpMM Strategy}
\begin{figure*}[t]
\begin{center}
\includegraphics[width=1\linewidth]{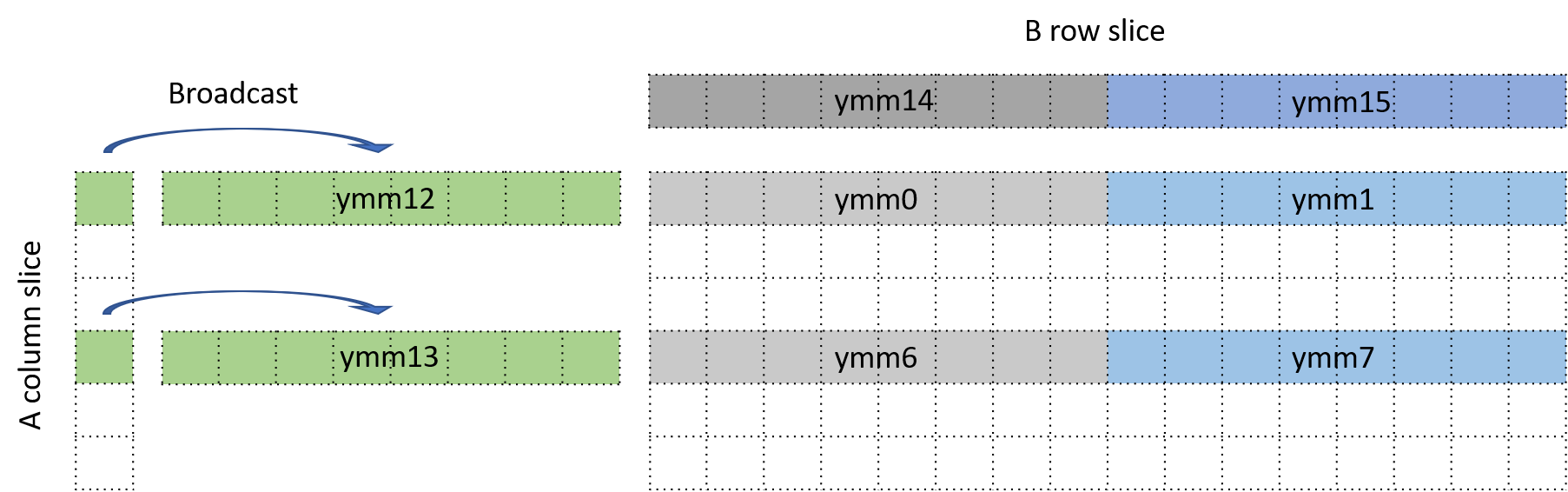}
\end{center}
   \caption{SpMM microkernel. Only the nonzero elements in an A column slice are processed. We iterate through the nonzeros in the A column slice, broadcasting each element into vector registers. Then the corresponding B row slice is loaded into vector registers. The vector registers holding A and B values are then multiplied and accumulated to the vector registers holding the output tile.}
\label{fig:spmm}
\end{figure*}

In SpMM, we adopt the same general strategy as dense matrix multiplication, as illustrated in Figure \ref{fig:spmm}. Similar to Libxsmm and SkimCaffe, we focus on optimizing a sparse microkernel \cite{georganas2019high,park2016faster}. Our sparse microkernel is based on the dense microkernel described above. Since A is sparse, only the nonzero elements in each A column slice needs to be processed. Since B is dense, the vector registers that load the B row slice are fully packed. For each A column slice, we iterate over the nonzeros. We broadcast each nonzero to a vector register, multiply it with the vector registers holding the B row slice, and accumulate the result to the output tile. Although the accumulation position depends on the random nonzero position, we can use registers for the output tile by statically encoding the accumulation position in the vector register name. This strategy is inspired by the code unrolling approach first described in \cite{wang2020sparsert}, and represents one crucial difference between our code generator and Intel's Libxsmm and SkimCaffe \cite{georganas2019high,park2016faster}. 

Similar to dense GeMM, we also employ tiling in SpMM. Since the nonzero locations in the A column slice can be random due to unstructured sparsity, we can expect random memory accesses into the dense B matrix. We tile the dense B matrix to optimize for CPU cache behavior along the reduction dimension at the cost of reloading the output tile from memory. This is commonly referred to as ``split-K'' in dense GeMM optimization literature \cite{kerr2017cutlass}. This optimization significantly improves performance on SpMM problems with large reduction dimensions.

Since the sparsity pattern is statically known, we can perform autotuning on the sparse weight matrix to find the best settings for the microkernel and the tiling factor. Our sparse code generator directly generates x86 assembly in contrast to the approaches in Libxsmm and SkimCaffe \cite{georganas2019high,park2016faster}. This was mainly done to accelerate the autotuning process by bypassing the compiler and improve register placement in the generated code.

Finally, we notice that our generated kernel traverses the sparse matrix in a Z-curve whose precise shape depends on the dimensions of the microkernel output tile and split-K tiling strategy. Typical sparse data formats such as CSR and CSC do not support efficient traversal of the sparse matrix in this manner. Since we statically know the exact order in which the sparse matrix values will be accessed depending on our microkernel parameters and tiling strategy, we reorganize the sparse matrix values such that they will be accessed sequentially. In effect, each sparse matrix is stored in its own custom format, optimized for the SpMM routine. This represents another difference between our approach and Libxsmm and SkimCaffe.

\subsubsection{Sparse Convolution}
We apply the direct convolution approach to implement sparse convolutions from our SpMM code generator \cite{park2016faster}. Briefly, the sparse convolution is treated as an SpMM where the sparse matrix is the filters treated as an $OC$ by $IC \times F_x \times F_y$ matrix. $OC$ and $IC$ are output and input channel counts and $F_x$ and $F_y$ are filter dimensions. The dense matrix is a virtual matrix that is realized on the fly from the input dense activations with the proper offsets \cite{park2016faster}. The SpMM is then optimized using the techniques described in the last subsection.

\subsection{Optimized Dense Kernels}
Dense operations in the neural network, such as pooling, self attention, and softmax, are mostly handled by the Intel oneDNN library. However, SparseDNN allows users to use other custom dense kernels. In this work, we use hand-optimized dense kernels for grouped convolutions. This is because the optimized kernels in oneDNN use a custom memory format for the input activations that is different from what our sparse kernels expect. 

In addition, the user could generate dense kernels from deep learning compiler frameworks such as TVM or Triton to use as plug-ins instead of oneDNN kernels if they provide better performance \cite{tillet2019triton,chen2018tvm}. We do not explore this option in this paper. 

\section{Results}
We evaluate SparseDNN in two different ways. We first evaluate our sparse code generator via kernel-level benchmarks on single operators. We then evaluate SparseDNN as a system with two end-to-end neural network inference benchmarks in CV and NLP. 

We consider two different CPU backends, both available via Amazon AWS: AMD Epyc 7R32 processors and Intel Xeon(R) Platinum 8124M processors. One key difference between the AMD processor and the Intel processor is that AMD hardware only supports AVX2 vector intrinsics while the Intel processor supports AVX512 instructions. We generate AVX2 code for AMD and AVX512 code for Intel processors. 

\subsection{Kernel Benchmarks}

\begin{figure*}[t]
\begin{center}
\includegraphics[width=1\linewidth]{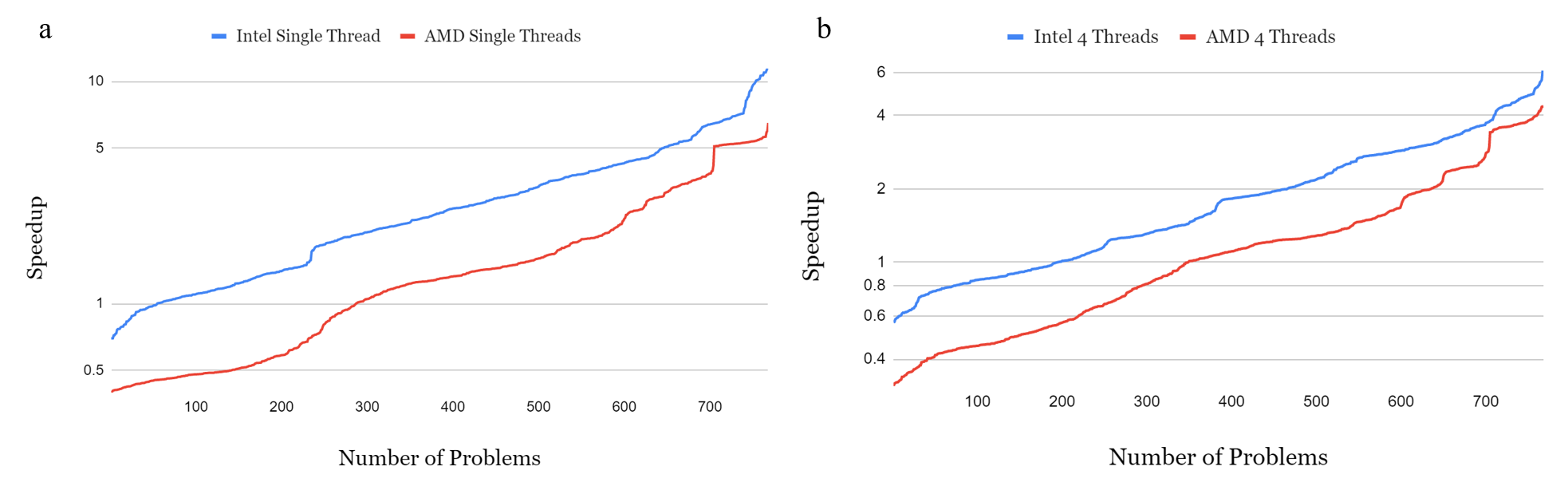}
\end{center}
   \caption{SpMM benchmark results for SparseDNN demonstrating speedup against Intel Libxsmm on a) one thread b) four threads. The plot shows the cumulative distribution function of speedups over the benchmark set of problems.}
\label{fig:spmm-results}
\end{figure*}
We evaluate our sparse code generator on two different operations, SpMM and sparse convolution, on a suite of typical problems encountered in deep learning. We consider both single-threaded performance and multi-threaded performance on 4 physical cores. As aforementioned, both scenarios are useful for end-to-end network inference. Single threaded performance is important for asynchronous inference, where each CPU handles a separate stream of inference requests, while multi-threaded performance is important for synchronous inference where all CPUs handle the same stream of inference requests.

We compare our SpMM routines against dense GeMM kernels from Intel MKL on Intel CPUs and OpenBLAS on AMD CPUs. We also compare against the optimized SpMM implementation from Libxsmm on both Intel and AMD CPUs \cite{georganas2019high,park2016faster} . We compare our sparse convolution routines against dense convolution kernels provided by Intel oneDNN on both Intel and AMD CPUs. We also compare against the optimized sparse convolution implementation in Intel SkimCaffe \cite{park2016faster}.

To test our SpMM routines, we first examine a benchmark suite of synthetic sparse matrices with sizes ranging from 256 by 256 to 2048 by 2048, with sparsity ratios at 70\%, 80\% and 90\% and 95\%. The sparsity pattern is generated by populating the matrices with random numbers drawn from a unit Gaussian distribution and then setting values below a certain threshold to zero. We find that the sparsity pattern generated closely resembles the ones obtained via magnitude pruning. The dense matrix has either 128, 512 or 2048 columns. This resulted in a total of 768 test matrices.

The benchmarking results are presented in Figure \ref{fig:spmm-results}. For single-threaded comparisons, we achieve a geometric mean of 2.5x speedup over the optimized SpMM routine in Libxsmm on Intel CPUs and 1.3x on AMD CPUs. We can see from Figure \ref{fig:spmm-results}a that for single-threaded benchmark, SparseDNN is faster than Libxsmm on most problems in the benchmark suite, reaching a speedup factor of 10x on the best performing problems on Intel CPU. 

The speedup achieved for multi-threaded performance shown in Figure \ref{fig:spmm-results}b was lower given the added cost of synchronization, which we do not improve. We achieve a geometric mean of 1.7x speedup over Libxsmm on Intel CPUs and breakeven on AMD CPUs. The speedup against dense BLAS libraries is dependent on the sparsity level, which will be elaborated in the next section. 

\begin{figure}[t]
\begin{center}
\includegraphics[width=1\linewidth]{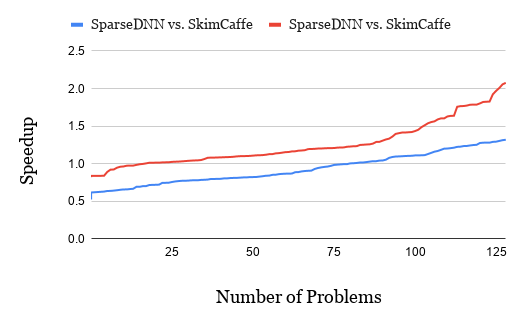}
\end{center}
   \caption{Single-threaded sparse convolution benchmark. Multi-threaded performance is similar.}
\label{fig:conv-results}
\end{figure}

To test our sparse convolution routines, we use a benchmark suite of sparse convolution problems with random sparsity generated in the same way as our SpMM benchmark. The number of input and output channels range from 32 to 256, while the image dimension is one of 7, 14, 28 and 56. The filter size is 3 by 3 with a padding size of 1. The problems are at either 90 or 95 percent sparsity. Compared to the sparse convolution kernels in SkimCaffe \cite{park2016faster}, we obtain a geometric mean 22\% speedup on Intel and 9\% slowdown on AMD single thread, as shown in Figure \ref{fig:conv-results}. Compared to the optimized dense convolution routines in oneDNN, we obtain a geometric mean speedup of 3.4x at 90\% sparsity and 6.3x at 95\% sparsity for AMD CPU. We obtain a geometric mean speedup of 3.0x at 90\% sparsity and 4.8x at 95\% sparsity for Intel CPU. The results for multi-threaded are similar. 

\subsection{Performance Analysis}
\begin{figure*}[t]
\begin{center}
\includegraphics[width=1\linewidth]{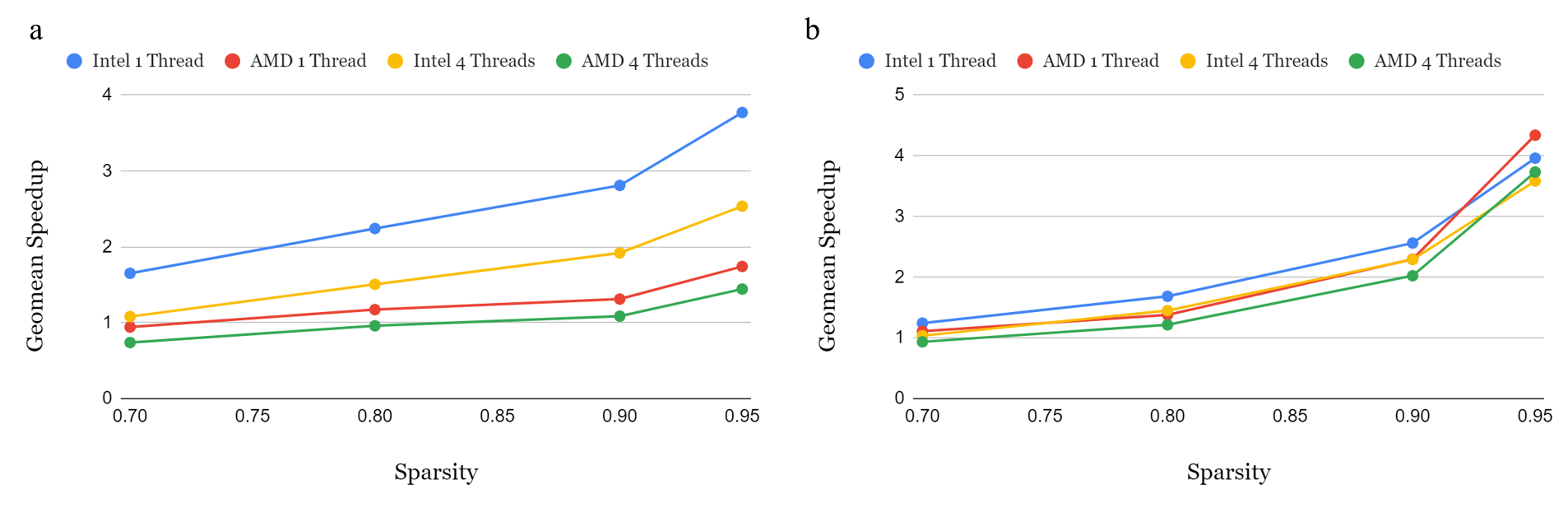}
\end{center}
   \caption{Geometric mean speedup obtained vs a) Libxsmm and b) dense BLAS kernel as a function of sparsity ratio. The matrices in the benchmark set had 70\%, 80\%, 90\% and 95\% sparsity. The break-even point is around 70\% in comparison to dense BLAS kernels, suggesting that sparse networks with more than 70\% zeros can be sped up using SparseDNN.}
\label{fig:sparse-curve}
\end{figure*}

\begin{figure}[t]
\begin{center}
\includegraphics[width=1\linewidth]{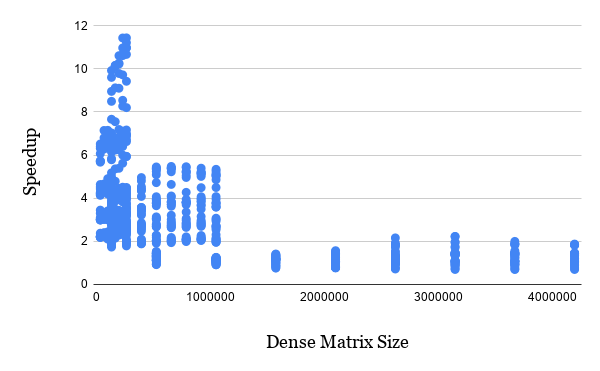}
\end{center}
   \caption{The speedup obtained against Libxsmm is strongly inversely correlated with the size of the dense matrix (r = -0.51).}
\label{fig:corr-size}
\end{figure}

This benchmark suite is aimed at answering two questions: 1) At what sparsity level can our sparse code generator provide advantages over Libxsmm and dense GeMM? 2) How does the performance change depending on the dimensions of the sparse and dense matrix operands?

Figure \ref{fig:sparse-curve} shows how the speedups achieved by SparseDNN with respect to Libxsmm SpMM routines and dense BLAS libraries scale with the sparsity level. From Figure \ref{fig:sparse-curve}a we can see that SparseDNN performs better with respect to Libxsmm with increasing sparsity, with a break-even point of around 70\% sparsity on Intel CPUs. SparseDNN performs worse on AMD CPUs, with a break-even point of around 80\% sparsity. 

Compared to dense BLAS libraries MKL and OpenBLAS in Figure \ref{fig:sparse-curve}b, we achieve a break-even point of around 70\% sparsity for both CPU architectures. This observation suggests that unstructured sparsity can translate into real speedups as long as the neural network has 70\% sparsity. At 95\% sparsity, we can achieve around 4x speedup over dense BLAS routines. 

Crucially, SparseDNN's break-even point with dense BLAS routines roughly coincides with the SparseDNN's break-even point with Libxsmm, which suggests that SparseDNN performs better for the vast majority of sparse problems in the regime where sparse computation is faster than dense computation. This shows the efficacy of incorporating the known sparsity pattern of the sparse matrix into static optimizations, which is the key difference between our approach and Libxsmm and SkimCaffe. 

We notice that the speedups we obtain relative to the dense BLAS libraries are consistent across both architectures while we perform worse against Libxsmm on AMD CPUs. This is attributed to the observation that the Intel Libxsmm library actually performs slightly better on the AMD CPU than on the Intel CPU tested, compared to the dense BLAS baselines. 

The most important factor in determining SparseDNN's performance is the dense matrix size. In Figure \ref{fig:corr-size}, we plot the single threaded speedup achieved by SparseDNN compared to Libxsmm on Intel CPUs against the size of the dense matrix. The results for AMD CPUs and multi-thread are similar. We see that while SparseDNN can achieve $>$ 10x speedups for small matrices, it is often on par with Libxsmm for very large matrices. 

As for sparse convolution, we find that we outperform SkimCaffe on problems where the number of output channels exceed the number of input channels, while we perform worse on problems where the number of output channels is small. We also perform relatively worse on small images since we do not register block across rows of the image as in SkimCaffe.

In summary, our sparse code generation technique outperforms state-of-the-art sparse libraries, and achieves several times speedup over dense BLAS libraries. This suggests that unstructured sparsity can be used efficiently to speed up neural network inference.

\subsection{End-to-end Benchmarks}
\begin{figure*}[t]
\begin{center}
\includegraphics[width=1\linewidth]{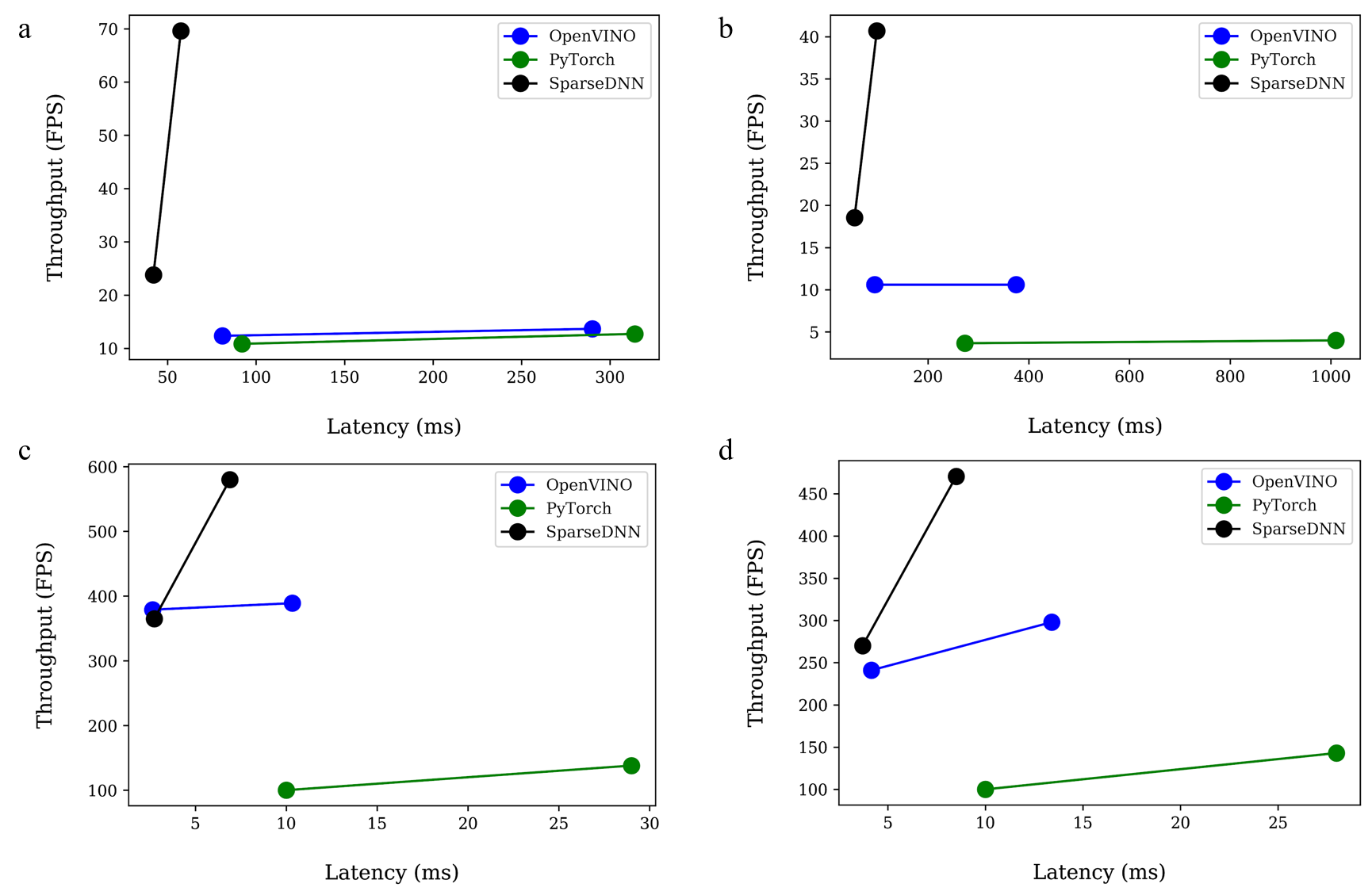}
\end{center}
   \caption{We show the latency/throughput tradeoff achieved by the three inference systems under synchronous and asynchronous inference. The dot corresponding to synchronous inference is always to the left of the dot corresponding to asynchronous inference. Ideally, we want low latency and high throughput (upper left corner of graph). a) Intel pruneBERT. b) AMD pruneBERT. c) Intel MobileNet V1. d) AMD MobileNet V1.}
\label{fig:network-results}
\end{figure*}

To evaluate the end-to-end inference performance of SparseDNN, we choose two different neural networks from CV and NLP that are representative of different operations popular in deep learning. The first benchmark is MobileNet V1, which is composed solely of grouped convolutions and 1x1 convolutions. We use a 90\% unstructured sparse model, where only the 1x1 convolutions are pruned \cite{elsen2019fast}. The grouped convolutions are dense. This model achieves 68.4\% top-1 accuracy on ImageNet, vs 70.9\% from an unpruned model.

The second benchmark is pruneBERT by Huggingface \cite{sanh2020movement}. It is a 95\% unstructured sparse BERT model, consisting mostly of fully connected layers and self attention units. Only the fully connected weights are pruned in this model. The self attention unit is dense. This model achieves an F1 score of 80 on the SQuAD v1.1 benchmark, vs. the unpruned model's score of 88. We benchmark only the encoder stack. We use 32-bit floating point for both models. 

We compare our end-to-end inference performance with two different inference frameworks. The first is the network implemented in Pytorch with dense operators, backed by the Intel MKL-DNN kernel library. Pytorch only provides experimental support for sparse operators. We found that the sparse operators are only faster than the dense equivalent under very high ($>$98\%) sparsity ratio, making them unsuitable for our benchmarks. 

The second is the Intel OpenVINO inference framework \cite{gorbachev2019openvino}. Intel OpenVINO is a highly optimized closed source framework for CPUs that perform complex optimizations such as layer fusion and intermediate activation format selection. It is typically faster than a naive Pytorch or Tensorflow implementation, even on AMD CPUs. However, it is less flexible than Pytorch due to the lack of an API to construct the network layer by layer. Instead, models must be converted to a supported IR format using the provided conversion scripts for different frontends such as Tensorflow 1.x and ONNX. While we were not able to convert the Huggingface pruneBERT model to OpenVINO IR, we were able to benchmark a supported BERT architecture that has the same encoder stack. 

Since this paper was first written, Neuralmagic Inc. has open-sourced a sparse neural network inference engine Deepsparse that is able to achieve impressive performance on a set of neural networks sparsified with another library SparseML. Deepsparse also optionally supports direct conversion from an ONNX model. Unfortunately the ONNX support falls short for both of the custom-pruned models used in this paper. Deepsparse also does not offer an operator-level API to support custom neural networks. Due to these difficulties, we don't include Deepsparse as a comparison point. 

For the three inference systems, SparseDNN, Pytorch and OpenVINO, we test the achieved latency and throughput in both synchronous and asynchronous execution. For both Intel and AMD CPUs, we assume we have access to four physical cores. We present our end-to-end results in Figure \ref{fig:network-results}. The result from synchronous inference (dot on the left) always has lower latency and lower throughput than the result from asynchronous inference (dot on the right). Results in the upper left corner of the graphs are more desirable as they correspond to higher throughput and lower latency. 

In all cases, Intel OpenVINO outperforms Pytorch. The performance gap is especially large for MobileNet V1, where OpenVINO employs data layout transformations and fuses some depthwise and groupwise convolution layers. 

We see that SparseDNN is able to outperform OpenVINO significantly in all cases except synchronous inference for MobileNet V1 on Intel CPUs. SparseDNN is particularly efficient in asynchronous inference, achieving a 5x improvement in throughput for pruneBERT on Intel CPUs with lower latency than \textit{synchronous} inference with OpenVINO. This suggests that pruneBERT inference with SparseDNN on a single core is faster than dense OpenVINO inference on four cores.

Although we can achieve even greater reductions in latency with synchronous inference, SparseDNN incurs a much steeper throughput penalty than OpenVINO or Pytorch. This reflects the earlier kernel-level benchmark results where SparseDNN does comparatively poorly in a multi-threaded scenario compared to single-thread. We are currently working on improving our OpenMP based threading implementation in SparseDNN to use an optimized custom thread pool implementation, as done in Amazon Sagemaker Neo \cite{liu2019optimizing}.

The performance gap between OpenVINO and SparseDNN decreases for MobileNet V1. This is due to the lower sparsity ratio (90\% vs. 95\%) and better performance of OpenVINO. While the sparse 1x1 contractions are greatly accelerated in SparseDNN, the dense depthwise convolution layers are handled better in OpenVINO. OpenVINO uses a custom data format that allows vectorization over the channels, and fuses some depthwise layers into the following 1x1 convolutions. We were not able to achieve end-to-end speedups using the oneDNN API for the depthwise layers. This prompted us to use hand-optimized depthwise convolution kernels, which resulted in a 50\% gain in throughput on Intel CPUs and a 58\% gain in throughput on AMD CPUs over OpenVINO. 

While the efficient sparse kernels are responsible for the bulk of the performance improvement, the network-level optimizations were also beneficial. Although on pruneBERT, these optimization only improved inference throughput by six and three percent respectively on Intel and AMD CPUs, they had a much greater effect on MobileNet V1. On Intel CPUs, they improved throughput by 27\% while on AMD CPUs, they improved throughput by 21\%. 

\section{Conclusion and Future Work}

In this work, we present SparseDNN, a deep learning inference engine that supports unstructured sparsity. At its core, SparseDNN relies on a sparse code generator that leverages the static sparsity patterns of the sparse weights to generate optimized kernels that are significantly faster than current state-of-the-art libraries. Network-level optimizations catering to sparsity such as buffer reuse and weights permutation improve the end-to-end inference results further, resulting in large latency and throughput gains over highly optimized dense inference engines. 

It is important to note that SparseDNN currently does not handle activation sparsity typically induced by activation functions such as relu. Activation sparsity is typically dynamic, which does not allow for the static optimization techniques leveraged by our code generator. In addition, its presence heavily depends on the activation function used, with newer activation functions such as gelu \cite{hendrycks2016gaussian} inducing much less sparsity than relu. 

Although currently applied to sparse inference, SparseDNN can also be applied to sparse training. This is because current sparse training algorithms typically use a fixed sparse network architecture or a fixed sparsity pattern for a number of iterations \cite{frankle2018lottery,evci2019rigging}. As a result, the sparsity pattern is in effect statically known, and the sparse kernels can be generated by SparseDNN before training or just-in-time every time the architecture changes. 

We note that in addition to network pruning, quantization is also a popular approach to compress neural networks. Recently, pruning has been successfully combined with quantization to produce sparse quantized networks \cite{kozlov2020neural}. We are currently working on adding support for those sparse low-precision operators.

This work adds to a growing corpus of literature that suggests unstructured sparsity is a viable option to obtain tangible speedups on commodity hardware \cite{gale2020sparse,elsen2019fast,park2016faster,wang2020sparsert,chen2019escoin}. We hope that by efficiently supporting unstructured sparsity in end-to-end network inference, SparseDNN can make it easier to productionize novel pruning algorithms and inspire future research into neural network compression.

\bibliographystyle{ACM-Reference-Format}

\bibliography{example_paper}



\end{document}